\documentclass{article}
\usepackage{graphicx} 
\usepackage{comment}

\usepackage{algorithmicx}
\usepackage{algpseudocode}
\usepackage{caption}

\usepackage{hyperref}

\usepackage{graphicx}
\usepackage{subcaption}
\usepackage{multirow}

\usepackage{tcolorbox}
\tcbuselibrary{breakable}

\usepackage[backend=biber,style=numeric,sorting=none]{biblatex}
\newtcolorbox{algo}[1][]{
  breakable,
  title=#1,
  colframe=black,
}

\title{Scalable Pretraining of Large Mixture of Experts Language Models on Aurora Super Computer}

\author{
\fontsize{11}{11}\selectfont Dharma Teja Vooturi, Dhiraj Kalamkar, Dipankar Das, Bharat Kaul \\ 
\fontsize{11}{11}\selectfont Parallel Computing Lab (India)\\
\fontsize{11}{11}\selectfont Intel Corporation \\ 
}

\date{}

\begin{document}

\maketitle

Abstract :  Pretraining Large Language Models (LLMs) from scratch requires massive amount of compute. Aurora super computer \cite{allcock2025auroraarchitectingargonnesexascale} is an ExaScale machine with 127,488 Intel PVC (Ponte Vechio) GPU tiles. In this work, we showcase LLM pretraining on Aurora at the scale of 1000s of GPU tiles. Towards this effort, we developed Optimus, an inhouse training library with support for standard large model training techniques. Using Optimus, we first pretrained  Mula-1B, a 1 Billion dense model and Mula-7B-A1B, a 7 Billion Mixture of Experts (MoE) model from scratch on 3072 GPU tiles for the full 4 trillion tokens of the OLMoE-mix-0924 dataset. We then demonstrated model scaling by pretraining three large MoE models Mula-20B-A2B, Mula-100B-A7B, and Mula-220B-A10B till 100 Billion tokens on the same dataset. On our largest model Mula-220B-A10B, we pushed the compute scaling from 384 to 12288 GPU tiles and observed scaling efficiency of around 90\% at 12288 GPU tiles. We significantly improved the runtime performance of MoE models using custom GPU kernels for expert computation, and a novel EP-Aware sharded optimizer resulting in training speedups up to 1.71x. As part of the Optimus library, we also developed a robust set of reliability and fault tolerant features to improve training stability and continuity at scale.

\section{Training Large Models}
Training a model with P parameters in BF16 mixed precision using the AdamW optimizer requires atleast 16P bytes of memory - 2P for weights, 2P for gradients, 4P for FP32 master weights, and 8P for FP32 optimizer states. As a result, training a small 7B model with Pytorch DDP is infeasible on PVC GPU tiles as 112 GB ($16\times 7$) memory requirement is beyond the 64 GB memory capacity available on a PVC GPU tile. To overcome this limitation and enable large model training, several techniques such as sharded data parallelism (e.g., ZeRO \cite{rajbhandari2020zero}, FSDP \cite{zhao2023pytorchfsdp}), model parallelism (Tensor \cite{shoeybi2019megatronlm} , Expert \cite{lepikhin2020gshard}, and Pipeline \cite{huang2019gpipe}), context parallelism \cite{liu2024ringattention}, activation checkpointing \cite{chen2016training} etc. have been proposed. In this section, we will go through the subset of the techniques we employed and specific implementations of them in our Optimus training library.
\\\\
\textbf{Sharded Optimizer (SO)} : In PyTorch DDP, the optimizer states are fully replicated on every data-parallel (DP) rank. After the backward pass, gradients are synchronized across ranks using an all-reduce operation. With the full gradient available locally, every rank independently updates its copy of the model weights.
In contrast, when using DP with a sharded optimizer, the optimizer states are partitioned (sharded) across DP ranks instead of being replicated. As a result, reduce-scatter operation is used to synchronize the gradients, distributing only the relevant gradient shards to each rank—specifically, the gradients corresponding to the parameters whose optimizer states it owns. Each rank updates only its assigned parameter shard. Finally, an all-gather operation is used to share the updated parameter shards so that all DP ranks once again hold a consistent and complete copy of the model. Sharding optimizer states allows us to train small MoE models with just DP.
\\\\
\textbf{Tensor Parallelism (TP)} : In TP, the model is partitioned horizontally. A decoder block in a dense transformer model has two modules : attention and MLP. In TP, the attention heads are divided in the attention module and intermediate size is divided in the MLP modules. To accumulate partial output activations and input gradients, we need to do allredue after/before attention and MLP in forward/backward pass on activations/gradients. Apart from decreasing the memory requirement for the GPU, TP allows us to scale the training to more number of GPUs without increasing the global batch size.
\\\\
\textbf{Expert Parallelism (EP)} : A decoder block in a Mixture-of-Experts (MoE) model has two modules : attention and SparseMoE. The SparseMoE module contains multiple experts along with a router. In Expert Parallelism (EP), the experts are divided across the EP ranks, while the non-expert parameters are replicated on each rank. Because expert parameters make up the majority of the model’s weights, distributing them allows us to train medium sized MoE models. In EP, global batch size scales with the number of GPUs just like DP.
\\\\
\textbf{Pipeline Parallelism (PP)} : In Pipeline Parallelism (PP), the model is partitioned vertically across layers. Each rank processes forward/backward for it's corresponding layer block and sends the activation/gradient to the next/previous rank. In PP, the input batch is divided into micro-batches and these are scheduled across PP ranks using different scheduling strategies. PP truly allows us to train arbitrary large models as the division is across layers.
\\\\
\textbf{Selective Activation Checkpointing (SAC)} : In forward pass, intermediate activations need to be stored to compute gradients in the backward pass. These intermediate activations can take up significant amount of memory. In SAC, for the selected blocks of the forward pass, only the input activation to the selected block is stored instead of all the activations in that block. During the backward pass for the selected block, the forward pass is recomputed from the stored input and then the backward pass is computed. 
\\\\
In our Optimus training library, 
\begin{itemize}
    
    \item We implemented two reference huggingface models \href{https://huggingface.co/allenai/OLMo-1B-hf}{allenai/OLMo-1B-hf} and \href{https://huggingface.co/allenai/OLMoE-1B-7B-0924}{allenai/OLMoE-1B-7B-0924}
     for the dense and MoE models respectively with model parallelism (Tensor, Expert and Pipeline Parallelism).

    \item We implemented gpipe \cite{huang2019gpipe}, 1f1b \cite{narayanan2019pipedream} , and interleaved-1f1b \cite{narayanan2021efficient} schedules for pipeline parallelism (PP) and enabled support for doing expert load balancing auxiliary loss for MoE model training with PP.

    \item We implemented selective activation checkpointing for three modules : norm layer, attention block, SparseMoE block with an option for selecting one or more modules during training.
    
    \item We developed a drop-in optimized FastSparseMoEBlock module that significantly speeds up MoE model training when compared to the SparseMoEBlock module in Hugging face OLMoE model. (See Section 3.1)

    \item We implemented sharded optimizer for AdamW and also developed EPSO, a novel expert parallelism (EP) aware sharded optimizer that results in better performance when EP is used for training. (See Section 3.2)
\end{itemize}

\section{Results}
We first pretrained Mula-1B dense and Mula-7B-A1B MoE models on the 4 trillion token OLMoE-Mix-0924 dataset \cite{olmoe2024}. We then demonstrate MoE model scaling with model sizes ranging from 20B to 220B parameters, and compute scaling from 384 to 12288 GPU tiles. Mula models follow the same model architecture as OLMo \cite{groeneveld-etal-2024-olmo} and OLMoE \cite{olmoe2024} for dense and MoE models respectively.

\begin{table}[ht]
    \centering
    \begin{tabular}{l|c|c|c|c|c}
         & Mula- & Mula- & Mula- & Mula- & Mula-  \\ 
          &  1B & 7B-A1B & 20B-A2B & 100B-A7B & 200B-A10B  \\ \hline
        Layers  & 16 & 16 & 32 & 48 & 64 \\
        Hidden size & 2048 & 2048 & 2048 & 3072 & 3072 \\
        Attention heads  & 16 & 16 & 16 & 24 & 24 \\
        Head size  & 128  & 128 & 128 & 128 & 128 \\ 
        Intermediate size  & 8192 & 1024 & 1024 & 1536 &  1536 \\ 
        Experts  & - & 64 & 96 & 144 & 240 \\
        Chosen Experts   & - & 8 & 8 & 8 & 8 \\ \hline
        Total parameters  & 1.3 B & 6.9 B & 20 B & 100 B & 220 B \\
        Active parameters & 1.3 B & 1.3 B & 2.4 B &  7.6 B & 10 B \\ 
        
    \end{tabular}
    \caption{Model configurations.}
    \label{tab:mula_model_configs}
\end{table}
\newpage
\subsection{Mula-1B and Mula-7B-A1B}
We pretrained Mula-1B and Mula-7B-A1B on 256 nodes (3072 GPU tiles) with DP=3072 and AdamW sharded optimizer. We closely follow the training recipe and hyper parameters from \cite{olmoe2024} with differences in context size (2048 instead of 4096), global batch size tokens (6.3 Million instead of 4 Million), and gradient reduction (bfloat16 instead of float32). The minimum and peak learning rates are set to 4e-5 and 4e-4 respectively. We do a linear warmup for the first 2500 steps and follow a cosine decay learning rate schedule. We choose weight\_decay value of 0.1 and apply on all the parameters. In AdamW optimizer, we chose (beta1=0.9, beta2=0.99, and eps=1e-8). We do gradient clipping with gradient norm of 1.0 and apply clipping only after the warmup steps.

\begin{figure}[h!]
    \centering
    
    \begin{subfigure}{0.49\textwidth}
        \centering
        \includegraphics[width=\linewidth]{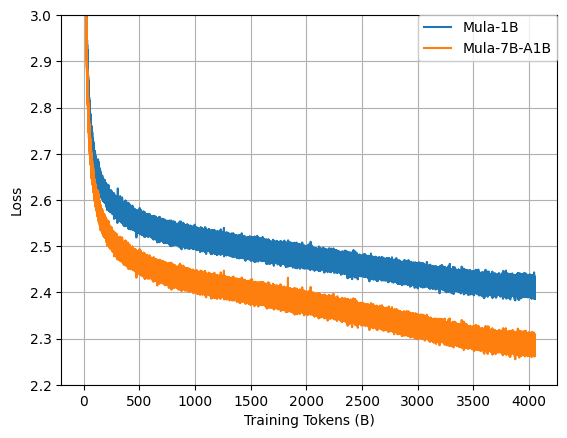}
        \caption{Mula-1B and Mula-7B-A1B }
        \label{fig:mula_loss_curve}
    \end{subfigure}
    \hfill
    \begin{subfigure}{0.49\textwidth}
        \centering
        \includegraphics[width=\linewidth]{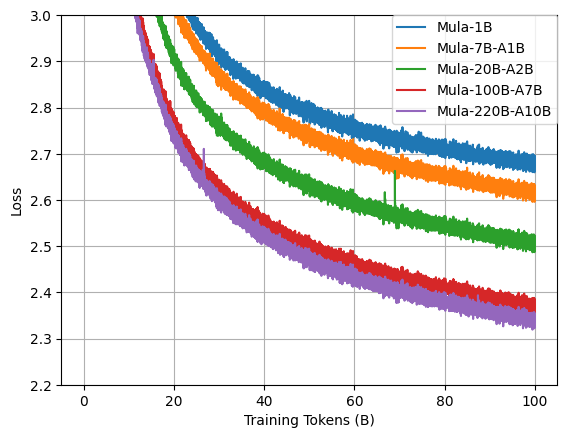}
        \caption{Model scaling}
        \label{fig:mula_large_loss_curve}
    \end{subfigure}
    \caption{(Left) Training loss on the 4 trillion token OLMoE-Mix-0924 dataset. (Right) Training loss till 100 B training tokens on the same dataset}
    \label{fig:both}
\end{figure}

\begin{figure}[h!]
    \centering
    
    \begin{subfigure}{0.49\textwidth}
        \centering
        \includegraphics[width=\linewidth]{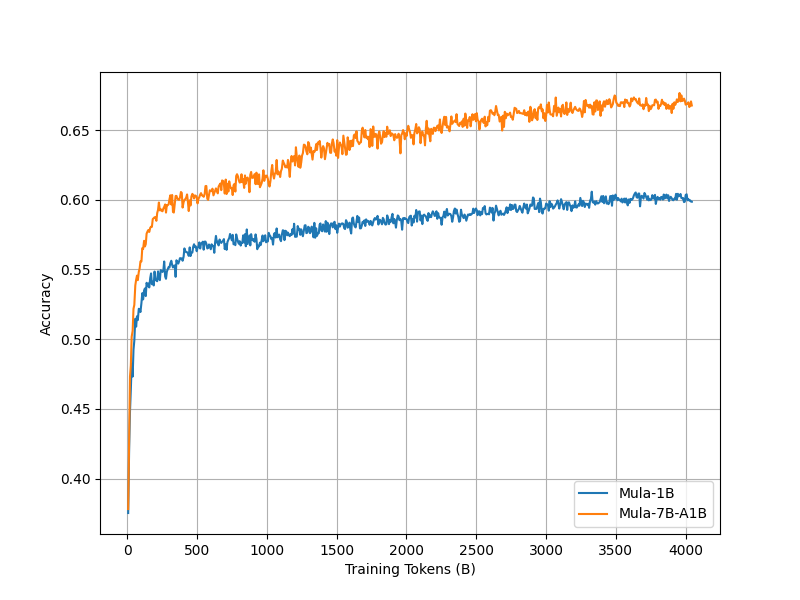}
        \caption{Average benchmark accuracy }
        \label{fig:average_benchmark}
    \end{subfigure}
    \hfill
    \begin{subfigure}{0.49\textwidth}
        \centering
        \includegraphics[width=\linewidth]{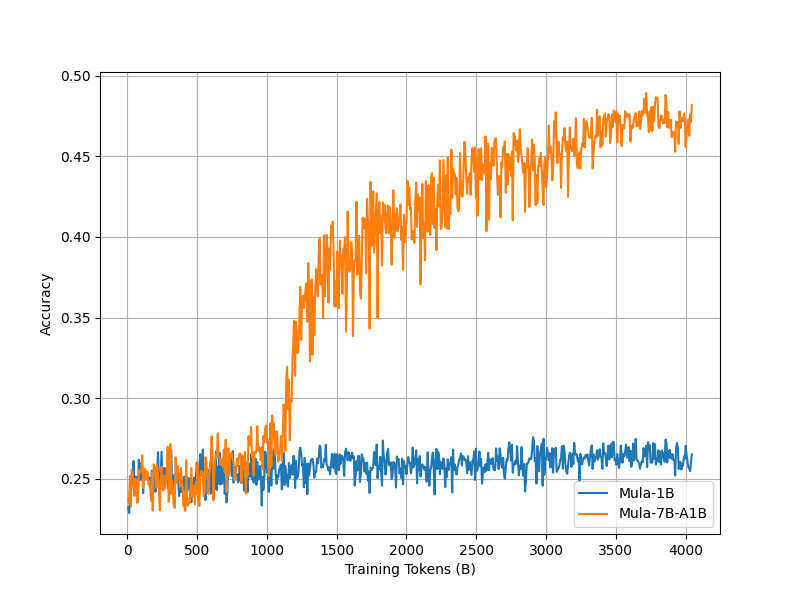}
        \caption{MMLU benchmark accuracy}
        \label{fig:mmlu_benchmark}
    \end{subfigure}
    
    \caption{Benchmark performance progression of Mula-1B and Mula-7B-A1B models on 4 trillion token OLMoE-Mix-0924 dataset.}
    \label{fig:mula_vs_mulae}
\end{figure}

\begin{figure}[h!]
    \centering
    \includegraphics[width=0.95\linewidth]{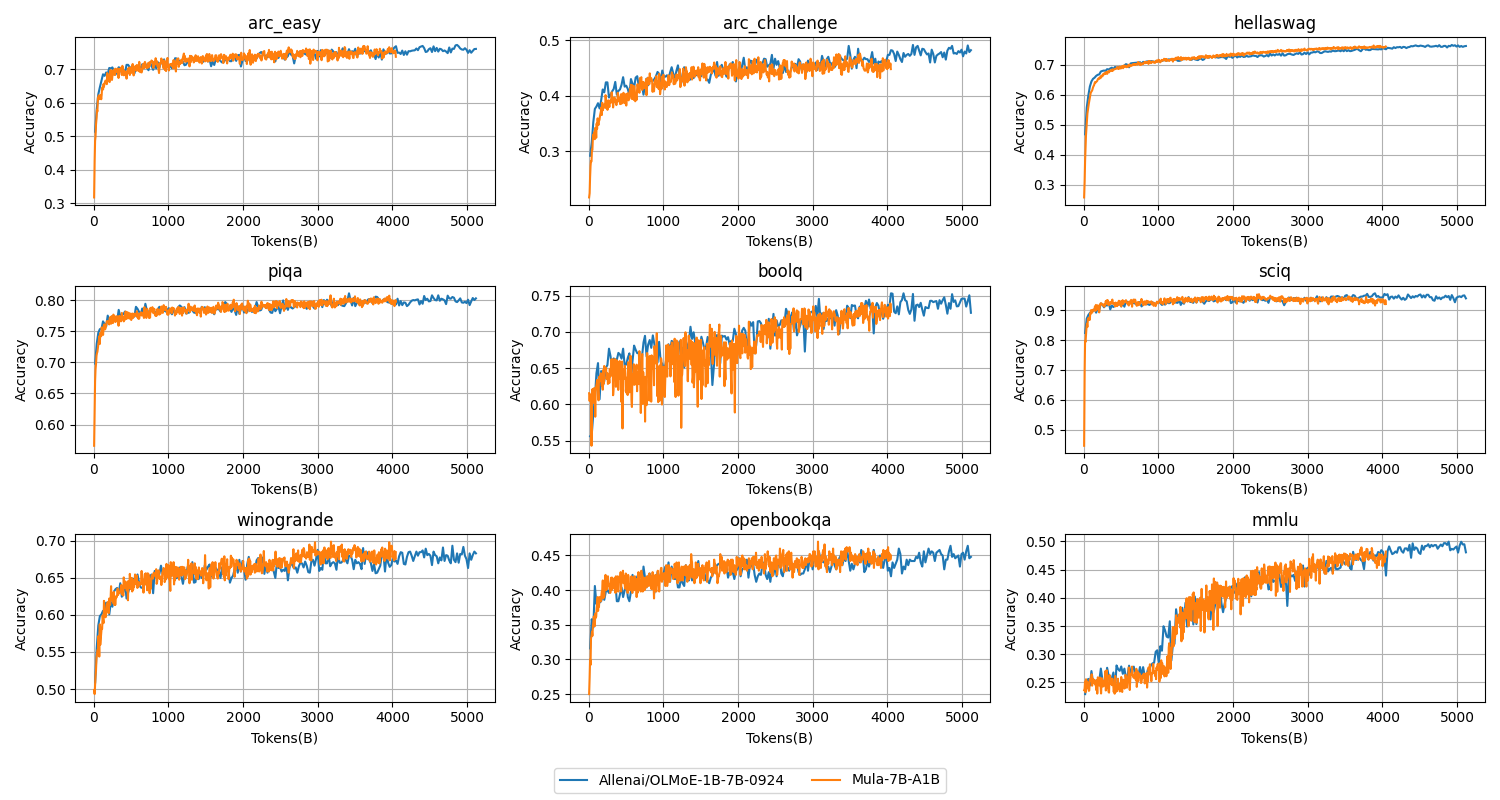}
    \caption{Mula-7B-A1B and allenai/OLMoE-1B-7B-0924 benchmark performance progression with evaluations done on intermediate training checkpoints.}
    \label{fig:intermediate_evaluations}
\end{figure}

Figure \ref{fig:mula_loss_curve} shows the training loss curves, and Table \ref{tab:benchmark_evaluation} shows the performance of the pretrained models on the standard benchmarks. Both models have the same compute but the Mula-7B-A1B achieves 6.7\% more accuracy than the Mula-1B model. The 6.7\% gap might seem less but from Figure \ref{fig:average_benchmark} we can see that the Mula-7B-1B model reaches Mula-1B accuracy at around 500 Billion tokens, which is just 1/8th of the training tokens. Out of all the benchmarks, mmlu is the toughest benchmark and from Figure \ref{fig:mmlu_benchmark} we can see that Mula-1B mmlu accuracy stays flat and is same as a random model. But for Mula-7B-1B it jumps around 1 trillion tokens and steadily increases. From these observations, we can say that at iso compute MoE models are more accurate than dense models. From Table \ref{tab:benchmark_evaluation}, we can also see that our Mula-7B-A1B model achieves similar performance as OLMoE-1B-7B-0924 \cite{olmoe2024} which is also trained on the OLMoE-Mix-0924 dataset for 1.3 epochs instead of 1 epoch. We evaluated the intermediate checkpoints released by AllenAI for OLMoE-1B-7B-0924 and from Figure \ref{fig:intermediate_evaluations} we can see that our model Mula-7B-A1B accuracy tracks well with the OLMoE-1B-7B-0924 model reinforcing the correctness of the software stack.

\begin{table}[h!]
    \centering
    \begin{tabular}{c||c|c||c}
    Benchmark & Mula-1B & Mula-7B-A1B & OLMoE-1B-7B-0924 \\
    \hline
    
arc\_easy             & 70.2  & 75.4  & 76.5 \\ \hline
arc\_challenge        & 39.6  & 45.5  & 48.5 \\ \hline
hellaswag            & 67.7  & 76.3  & 77.0 \\ \hline
piqa                 & 76.5  & 79.1  & 80.7 \\ \hline
boolq                & 66.2  & 73.1  & 74.5 \\ \hline
sciq                 & 93.7  & 93.4  & 94.6 \\ \hline
winogrande           & 62.7  & 67.2  & 69.0 \\ \hline
openbookqa           & 39.8  & 45.6  & 44.8 \\ \hline
mmlu                 & 25.9  & 47.3  & 50.5 \\ \hline
Average              & 60.3  & 67.0  & 68.5 \\

    \end{tabular}
    \caption{Benchmark performance of Mula-1B dense , Mula-7B-A1B MoE and allenai/OLMoE-1B-7B-0924 MoE models.}
    \label{tab:benchmark_evaluation}
\end{table}

\subsection{Model Scaling} 
Mula-7B-A1B model is chosen as the base model from which we generate three large MoE models : Mula-20B-A2B, Mula-100B-A7B, and Mula-220B-A10B (See Table \ref{tab:mula_model_configs} for model configuration details). We trained all the models on 256 nodes (3072 GPU tiles) with the same training recipe used for training Mula-1B and Mula-7B-A1B. Mula-20B-A2B is a 20 Billion parameter model with 2.4 Billion active parameters trained with 12-way expert parallelism (EP=12) within the node of 12 GPU tiles. Mula-100B-A7B is a 100 Billion parameter model with 7.6 Billion activate parameters trained with 4-way pipeline parallelism (PP=4) across nodes and EP=12 within the node. For Mula-100B-A7B, we used activation checkpointing for the expert blocks. Mula-220B-A10B is a 220 Billion parameter model with 10 Billion active parameters trained with PP=8 across nodes and EP=12 within the node. For Mula-220B-A10B, we used activation checkpointing for norms, attention and expert blocks. For PP, we used 1f1b for micro batch scheduling. For sharded optimizer, we used EPSO (See Section 3.2). From Figure \ref{fig:mula_large_loss_curve}, we can see the expected behavior of training loss being indirectly proportional to the model size. As our primary objective is to demonstrate model scaling with different model parallel configurations, we trained the models only till 100 Billion tokens.

\subsection{Compute Scaling}
\begin{figure}[]
    \centering
    
    \begin{subfigure}{0.44\textwidth}
        \centering
        \includegraphics[width=\linewidth]{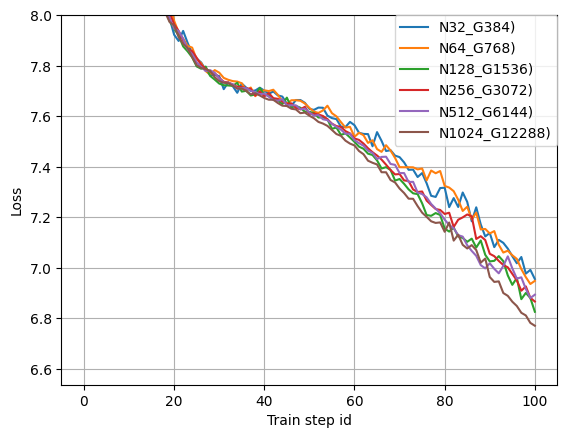}
        \caption{Training loss }
        \label{fig:compute_scaling_training_loss-step}
    \end{subfigure}
    \hfill
    \begin{subfigure}{0.54\textwidth}
        \centering
        \includegraphics[width=\linewidth]{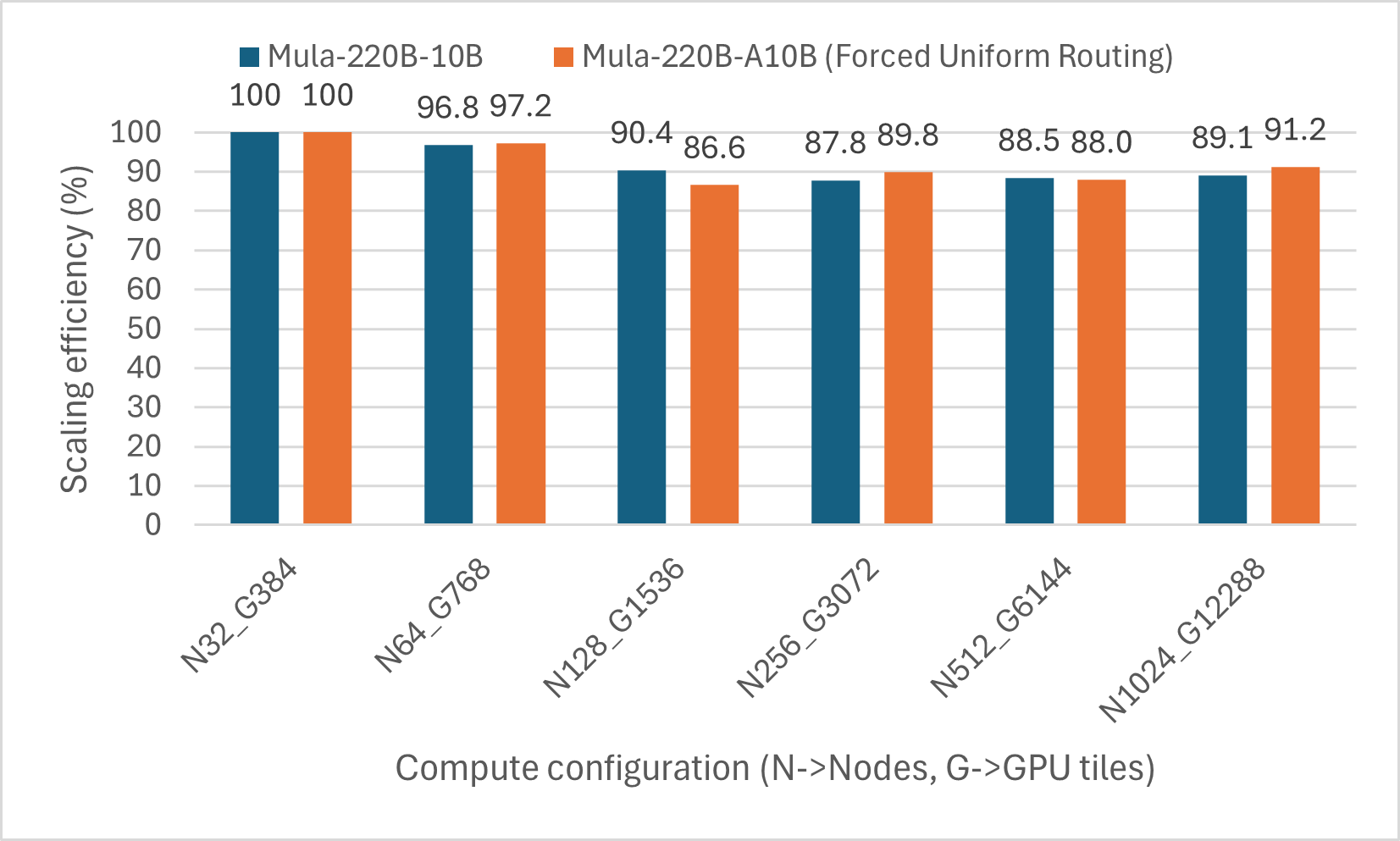}
        \caption{Scaling efficiency}
        \label{fig:mula_scaling_efficiency}
    \end{subfigure}
    
    \caption{Compute scaling of Mula-220B-10B model pretraining from 32 nodes (768 GPU tiles) to 1024 nodes (12288 GPU tiles)}
    \label{fig:compute_scaling}
\end{figure}

We chose our largest model Mula-220B-A10B and scaled the compute from 32 nodes (384 GPU tiles) to 1024 nodes (12288 GPU tiles). From Figure \ref{fig:compute_scaling_training_loss-step}, we can see that there is a healthy decrease in training loss in general and a trend of loss value being inverse proportional to the compute scaling. This trend is expected because with scaling, the batch size increases and higher batch size leads to better gradient approximation and model weight updates leading to a lower loss value. From Figure \ref{fig:mula_scaling_efficiency}, we can see that there is only a 3\% drop in scaling efficiency going from 384 to 768 GPU tiles. But going to more than thousand GPU tiles, the scaling efficiency drops by 10\% and stays around 90\% from 1538 to 12288 GPU tiles. Unlike dense models which have uniform computation across GPUs and training steps, MoE models have irregular compute across GPUs and training steps due to non-uniformity in expert selection. To rule out the effects of that on scaling data, we did the same training runs with FUR (Forced Uniform Routing), where all the experts receive the same number of tokens in the same pattern which ensures uniformity across GPUs and training steps. From Figure \ref{fig:mula_scaling_efficiency}, we can see that even with FUR, we can can see similar scaling efficiency dynamics as the regular training runs.

\section{MoE Performance Optimizations}
An MoE model is more complex in compute when compared to dense model. Due to this, off the shelf implementations from Hugging face are sub optimal. In this section, we will go through the two performance optimizations we developed to speedup MoE model training. One is FastSparseMoE, which is an optimized dropin module for the SparseMoE module in OLMoE hugging face implementation. Second is the EP-Aware sharded optimizer which exploits EP semantics and offers better performance over standard sharded optimizer.

\begin{table}[h]
\centering
    \begin{tabular}{l|c|c|c|c|c}
          & \multicolumn{2}{c|}{FastSparseMoE } & \multicolumn{2}{c|}{EP-Aware Sharded} & FSMOE+ \\
         & \multicolumn{2}{c|}{(FSMOE)} & \multicolumn{2}{c|}{Optimizer (EPSO)} & EPSO \\ \hline
        Model & F+B & Training & Optimizer & Training  & Training \\ \hline
        Mula-7B-A1B     & 2.83 & 1.71 & - & - & 1.71 \\
        Mula-20B-A2B    & 1.33 & 1.11 & 1.36 & 1.19 & 1.35 \\
        Mula-100B-A7B   & 1.51 & 1.31 & 1.23 & 1.06 & 1.41 \\
        Mula-220B-A10B  & 1.66 & 1.48 & 1.07 & 1.01 & 1.51
    \end{tabular}
    \caption{Speedup from MoE performance optimizations at the component and full training level. FSMOE effects forward (F) and backward (B) components of training. EPSO effects only optimizer component of the training with EP.}
    \label{tab:moe_perf_speedup}
\end{table}

\subsection{FastSparseMoE}
\begin{figure}[h]
    \centering
    \includegraphics[width=0.85\textwidth]{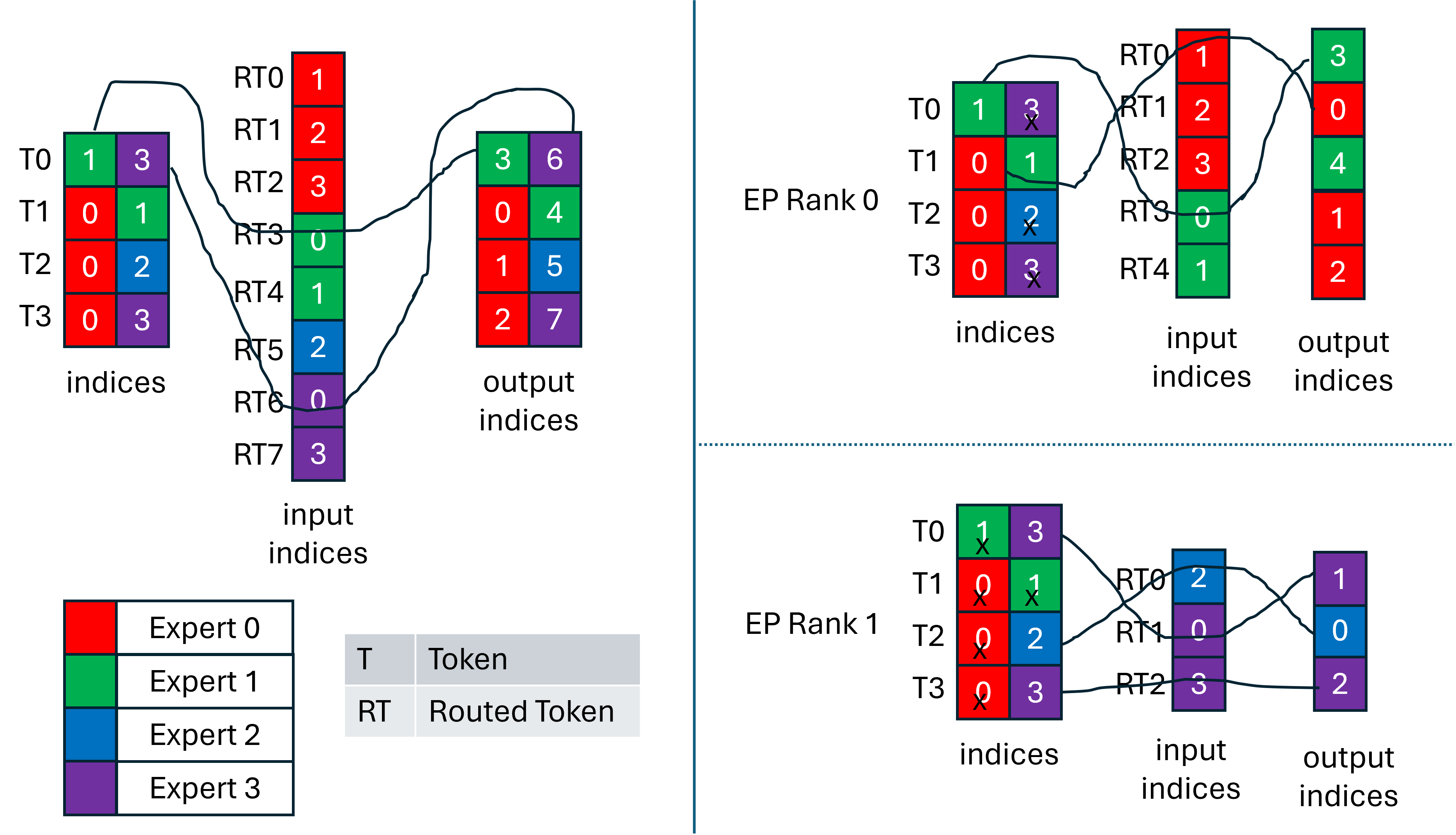}
    \caption{Index generation (input\_indices and output\_indices) example on four input tokens (T=4) with four experts(N=4) and two chosen experts per input token (K=2). In the EP case on the right with two ranks, experts 0 and 1 are placed on rank 0, and experts 2 and 3 are placed on rank 1.}
    \label{fig:sample}
\end{figure}

SparseMoEBlock in an MoE transformer model consists of N experts and a router. When Expert Parallelism (EP) is applied on an MoE model, the experts and the router in the SparseMoEBlock are divided and replicated among the EP ranks respectively, which results in each rank having N/EP experts and the router. Algorithm 1 describes our five staged optimized implementation on the GPU for FastSparseMoEBlock with EP. 
\\
\textbf{Stage 1 (Token Communication)} The input to the SparseMoEBlock is first sent to the router, a Linear layer that takes S input tokens and produce logits of size [S,N]. Softmax is then applied on the logits and TopK is applied on the softmax output to choose K experts for each input token. The chosen experts may not be there on the same rank, so the input tokens need to be communicated to the appropriate ranks. This communication is non-uniform and irregular as an input token need to be communicated only with the ranks where it's chosen experts reside. Ideally, all2all operation should be used to perform this communication. But we observed that sending all input tokens of an EP rank to all other EP ranks using allgather operation is faster despite having more communication volume. This can be attributed to efficient implementation of allgather in OneCCL library due to it's regular and uniform communication pattern. In lines 11-13, we communicate the input and the TopK outputs (weights and indices) among the EP ranks. In the forward pass, we do allgather on the input, weights, and indices. In the backward pass, we apply reducescatter on the input and weights to accumulate partial gradients.
\\\\
\textbf{Stage 2 (Token Counting)} : Each rank has to calculate two types of counts: 1) token\_counts : The number of input tokens routed to an expert in that rank. 2) expert\_counts : The number of selected experts that are local in that rank. Lines 25-37 shows our GPU kernel that does the token counting. The input to the kernel is the indices tensor which has the K selected expert ids of all input tokens from all the EP ranks. We launch TH threads and each thread is mapped to process a row block of the indices tensor. Each thread scans it's row block and only processes experts that are local to that rank (Condition in Line 29). If a selected expert is local to the rank, the thread increments the corresponding token and expert count by 1. In Lines 39-43, we sum up the partial token counts from all threads to get final counts and also perform prefix sum operations to calculate boundary information needed for the subsequent stages.
\\\\
\textbf{Stage 3 (Index Generation)} : Each rank calculates two types of indices: 1) input\_indices : Indices of input tokens selected for the expert in that rank.  2) output\_indices : Indices of input\_indices for the selected expert in that rank. Lines 53-72 shows our GPU kernel that does the index generation. The input to the kernel are the indices, token and expert counts we computed in Stage 2. The threading, thread mapping to indices tensor, and filtering the local experts remains the same as the token counting kernel in Stage 2. Additionally, we initialize counter tensor which keeps track of how many tokens are routed for a given expert and thread. If a selected expert is local to the rank, the thread first gets the base value from token\_counts (Line 61) and offset value from counter (Line 62). The input token id is then set at (base+offset) index in input\_indices, and the output\_indices stores (base+offset) value. Figure \ref{fig:sample} shows an example of index generation with and without EP. 
\\\\\\

\textbf{Stage 4 (Expert Computation)} : The input to each expert is a subset of input tokens for which that expert is chosen. In line 74, we use the input\_indices that is generated from Stage-3 to prepare input for the experts. An expert is nothing but an MLP block with three Linear layers : gate\_proj, up\_proj, and down\_proj. There is no dependency between computation of experts and they can be processed together to improve the efficiency by running multiple GEMMs with a single kernel as opposed to running them separately. This is especially useful when the GEMMs are small and experts are many. In our implementation, we merge the weights of a Linear layer across experts in a rank into a single weight tensor. For example, gate\_weight in Line 76 is obtained by concatenating gate\_proj.weight tensor from all the experts in that rank. This merged weight along with merged input and boundary information from cum\_token\_counts allows us to use Grouped\_mm operations as shown in Lines 76,77, and 79. 
\\\\
\textbf{Stage 5 (Output Reduction)} : An input token is routed to K experts and the K expert outputs have to weighted averaged using weights (Output of TopK) to produce the final output. In EP, the experts are divided across the EP ranks, so the K selected experts are also distributed the across EP ranks $(K=K_{1}+..+K_{EP})$. Because of this rank $r$ has to do weighted average for $K_r$ local experts to generate partial output. We implement this stage using two kernels : one for the forward (Lines 84-94) and other for the backward (Lines 101-111). For the forward kernel, we launch TxH threads with each thread mapping to a single output element. Each thread goes through the corresponding local selected experts (Line 88) and does the scales the expert output (Line 91) using the index information calculated in Stage 2. In Line 116, we do reducescatter to reduce the output and split the output appropriately. For the backward pass, we do allgather on the gradients to ensure all the ranks have the full output gradient. For the backward kernel, we launch $RT$ threads and each thread computes the gradient for the mlp\_output and weight by scaling and reduction respectively.
\\\\
The performance benefit from FastSparseMoE (FSMOE) optimization is shown in Table \ref{tab:moe_perf_speedup}. A training step is broken down into three components : forward, backward, and optimizer. FSMOE only effects forward and backward components. We achieve 1.33x to 2.83x speedup for the combined forward and backward components. At the end to end training level, we achieve 1.11x to 1.71x speedup. Mula-7B-A1B has the highest speedup because it is trained without EP and hence does not have the overhead of EP communication. 

\begin{algo}[Algorithm 1: FastSparseMoEBlock with Expert Parallelism(EP) ]
\begin{algorithmic}[1]
\State T = EP * S \Comment{S - Sequence length, T - Number of Tokens}
\State NR = N/EP \Comment{NR - Number of experts per rank}
\State n\_start = r * NR \Comment{Expert start index of EP rank r}
\State n\_end = (r + 1) * NR - 1 \Comment{Expert end index of EP rank r}
\State
\State \textbf{Stage 1 : Token Communication}
\State logits = Router(input) \Comment{Shape(input)=[S,H]}
\State probs = Softmax(logits) \Comment{Shape(logits)=[S,N]}
\State weights, indices = TopK(probs) \Comment{Shape(weights/indices)= [S,K]}
\State
\State input =  \textit{forward\_allgather\_backward\_reducescatter}(input)
\State weights =  \textit{forward\_allgather\_backward\_reducescatter}(weights)
\State indices =  \textit{forward\_allgather}(indices)

\State 
\State \textbf{Stage 2 : Token counting}
\State TBS = 8 \Comment{TBS - Token Block Size}
\State TH = T/TBS \Comment{TH - Number of Threads}
\State partial\_token\_counts = Zeros(NR*TH)
\State partial\_cum\_token\_counts = Zeros(NR*TH+1)
\State cum\_token\_counts =  Zeros(NR+1)
\State expert\_counts = Zeros(T)
\State cum\_expert\_counts = Zeros(T+1)

\State
\State \textbf{Kernel : compute\_partial\_token\_counts}
\For{$tid$ in $range(TH)$} \Comment{Mapped to GPU threads}
    \For{$i$ in $range(TBS)$}
        \For {$k$ in $range(K)$}
            \State n = indices[t,k]
            \If{(n\_start $\leq$ n $\leq$ n\_end)} 
                \State t = tid*TBS + i
                \State ln = n - n\_start
                \State partial\_token\_counts[ln*TH+t] += 1
                \State expert\_counts[t] += 1 
            \EndIf        
        \EndFor 
    \EndFor
\EndFor

\State
\State partial\_cum\_token\_counts[1:] = PrefixSum(partial\_token\_counts)]
\State cum\_expert\_counts[1:] = PrefixSum(expert\_counts)]
\For{$n$ in $range(NR+1)$}
\State cum\_token\_counts[n] = partial\_cum\_token\_counts[n*TH]
\EndFor

\State
\State \textbf{Stage 3 : Index generation}
\State RT = cum\_token\_counts[-1] \Comment{Routed tokens for experts on EP rank}
\State selected\_experts\_indices = Zeros(RT)
\State input\_indices = Zeros(RT)
\State output\_indices = Zeros(RT)
\State counter = Zeros(NR,TH)
\State
\State \textbf{Kernel : generate\_input\_and\_output\_indices}
\For{$tid$ in $range(TH)$} \Comment{Iteration mapped to a thread on GPU}
    \For{$i$ in $range(TBS)$}
        \State t = tid*TBS + i
        \State o\_ind = cum\_expert\_counts[t] 
        \For {$k$ in $range(K)$}
        \State n = indices[t,k]
        \If{(n\_start $\leq$ n $\leq$ n\_end)} 
            \State ln = n - n\_start
            \State base = partial\_cum\_token\_counts[ln*TH + tid]
            \State offset = counter[ln][tid]
            \State i\_ind = base + offset
            \State input\_indices[i\_ind] = t
            \State output\_indices[o\_ind] = i\_ind
            \State selected\_expert\_indices[o\_ind] = k
            \State counter[ln][tid] += 1
            \State o\_ind += 1
        \EndIf
        \EndFor    
    \EndFor
\EndFor

\State
\State \textbf{Stage 4 : Expert Computation}
\State mlp\_in = input[input\_indices]
\State gate\_out = Grouped\_mm(mlp\_in, gate\_weight, cum\_token\_counts)
\State up\_out = Grouped\_mm(mlp\_in, up\_weight, cum\_token\_counts)
\State mul\_out = Silu(gate\_out) * up\_out
\State mlp\_out = Grouped\_mm(mul\_out, down\_weight, cum\_token\_counts)

\State
\State \textbf{Stage 5 : Output Reduction}
\Function{ExpertOutputReductionForward}{ }
    \State output = Zeros(T,H)
    \For{$(t,h)$ in $range(T)Xrange(H)$} \Comment{Mapped to GPU threads}
    \State acc = 0
    \State base = cum\_expert\_counts[t]
    \State size = cum\_expert\_counts[t+1] - cum\_expert\_counts[t]
    \For{$i$ in $range(size)$}
        \State k = selected\_expert\_indices[base+i]
        \State index = output\_indices[base+i]
        \State acc += weights[t][k] * mlp\_out[index][h]
    \EndFor
    \State output[t][h] = acc
    \EndFor
    \State \Return output
\EndFunction
\State
\Function{ExpertOutputReductionBackward}{output\_grad}
    \State mlp\_out\_grad = Zeros(RT,H)
    \State weights\_grad = Zeros(EP*S,K)
    
    \For{$rt$ in $range(RT)$} \Comment{Mapped to GPU threads}
    \State o\_ind = output\_indices[rt]
    \State t = input\_indices[index]
    \State k = selected\_expert\_indices[rt]
    \State weight\_grad\_acc = 0
    \For{$h$ in $range(H)$}
        \State mlp\_out\_grad[o\_ind][h] = weights[t][k] * output\_grad[t][h]
        \State weight\_grad\_acc += mlp\_out[o\_ind][h] * output\_grad[t][h]
    \EndFor
    \State weights\_grad[t][k] = weight\_grad\_acc
    \EndFor
    \State \Return mlp\_out\_grad, weights\_grad
\EndFunction

\State 
\State output = ExpertOutputReduction(mlp\_out, weights)

\State output = \textit{forward\_reducescatter\_backward\_allgather}(output)
\State \Return output
\end{algorithmic}
\end{algo}

\subsection{EP-Aware Sharded Optimizer}
\begin{figure}
    \centering
    \includegraphics[width=0.8\linewidth]{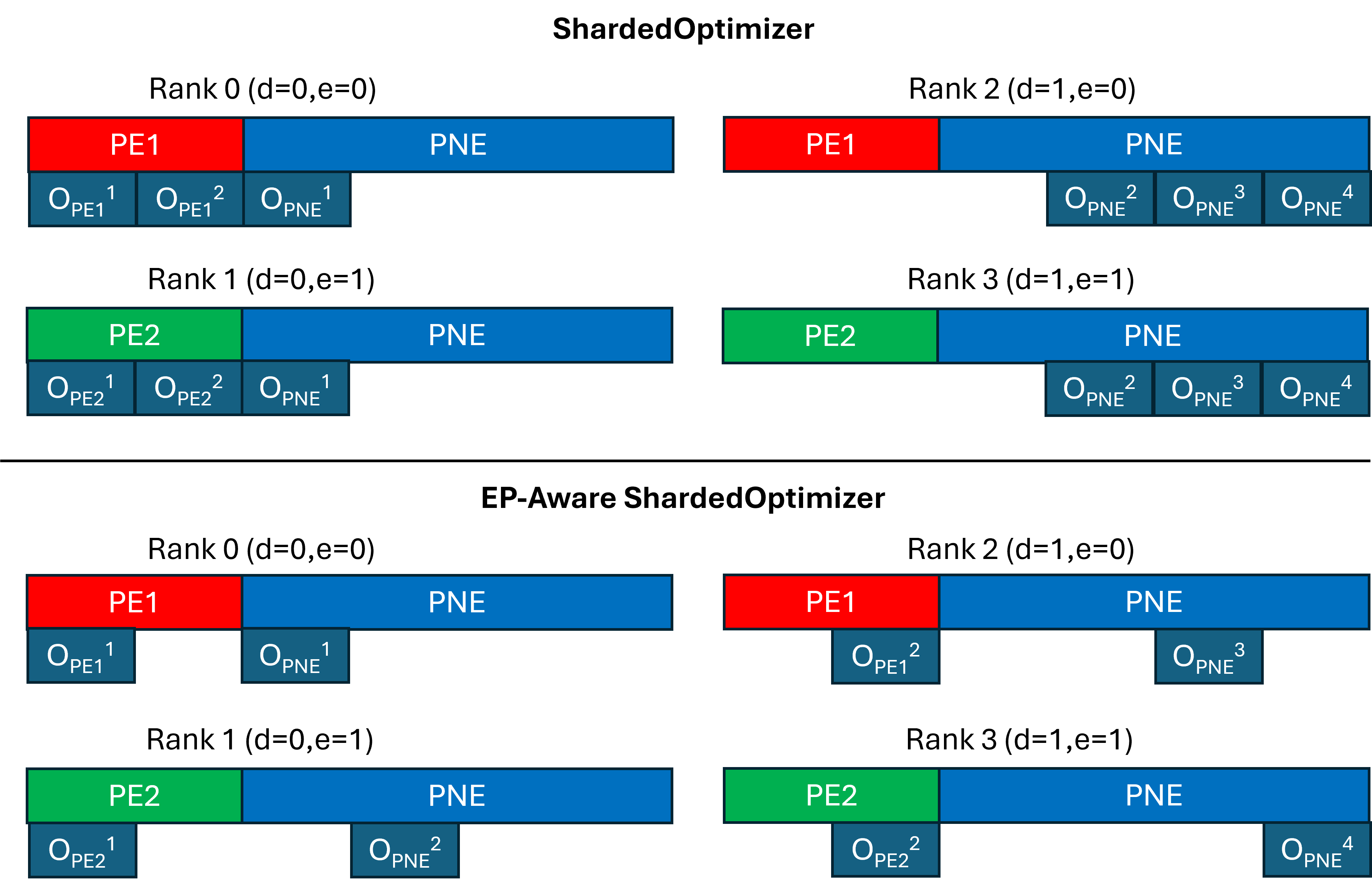}
    \caption{EP-Aware ShardedOptimier for MoE model with EP=2 and DP=2. Parameters $P = [PE, PNE]$. Expert parameters $PE = [PE1, PE2]$. In EP=2, PE1 and PE2 are placed on different expert ranks and PNE is replicated. } 
    \label{fig:epso_example}
\end{figure}

Data Parallelism (DP) is used to decrease training time by dividing the input batch and replicating the model across the DP ranks. Standard implementation of DP such as Pytorch DDP(Distributed Data Parallelism) also replicates optimizer states across DP ranks, which takes up a significant amount of memory. Sharded Optimizers (SO) address this problem by sharding the optimizer states across the DP ranks. Expert Parallelism (EP) divides the expert parameters in the MLP block  and replicates non-expert parameters(attention, embedding, lmhead, norms) across the EP ranks. When DP is applied on top on EP with SO, optimizer states corresponding to the non-expert parameters are replicated EP times because SO divides the optimizer states only across DP ranks. To overcome this issue, we propose Expert Parallel Aware Sharded Optimizer (EPSO). In EPSO, we first divide $P_r$ (parameters of rank $r$) into two groups $P_{r}^{E}$ and $P_{r}^{NE}$ based on how a parameter is replicated. Parameters in $P_r^E$ and $P_r^{NE}$ are replicated across DP and DPxEP ranks respectively. In DP+EP for MoE, $P_r^{E}$ and $P_r^{NE}$ contains parameters corresponding to experts and non-expert parameters respectively.  In EPSO, we follow a fine grained optimizer sharding strategy, where optimizer states of $P_r^E$ are sharded only across DP dimension and optimizer states of $P_r^{NE}$ are sharded across both DP and EP dimensions. Figure \ref{fig:epso_example} shows an example comparing EPSO with standard sharded optimizer.

The performance benefit from EPSO optimization is shown in Table \ref{tab:moe_perf_speedup}. EPSO only effects optimizer step and does not effect forward and backward passes in a training step. We achieve 1.07x to 1.36x speedup at the optimizer level. At the end to end training, we achieve speedup up to 1.19x.

\section{Reliability and Fault Tolerance}
Training at the scale of 1000s of GPUs puts pressure on all the components of the training infrastructure where instability and failure rate increases with scale. So there is a need to develop a robust set of reliability and fault tolerant features to ease the pressure on the infrastructure, ensure stable training, and recover quickly from failures. In this section, we will go through the features we developed that tackle issues related to data loading, model loading, model checkpointing, node failures etc.
\\\\
\textbf{Data preprocessing} : A typical hugging face dataset consists of data files, where each data file contains documents used for training. We perform data preprocessing to avoid tokenization during training, and to avoid random and non-contiguous memory access of the training data across many files. Our data preprocessing pipeline consists of three steps : Tokenization, Shuffling, and Sharding. In the tokenization step, we generate a token array Ti corresponding to the data file Di by tokenizing individual documents in Di and concatenating them with EOS token. Given a context size C, data file Di will have Ni training instances (Ni = Ti/C) of size C. In the shuffling step, we generate a permutation order P of size N, where N is the total number of training instances across the whole dataset (N=sum(Ni)). In the sharding step, we gather training instances from the tokenized data files based on the permutation order P and shard it into multiple numpy array files which are then loaded in mmap mode in a lazy manner. Our data processing pipeline allows all the data parallel ranks to load memory from a single file in a contiguous manner resulting in the bare minimal overhead for consuming tokens in the training.
\\\\
\textbf{Model Broadcasting} : In data parallelism (DP), parameters are replicated across the DP rank. So all the DP parallel ranks need to load the same model (initialized model at the beginning of the training or checkpointed model during the training). As all the DP parallel ranks load the model in parallel, it puts a huge pressure on the filesystem and can cause hangs in model loading itself. To avoid this, we only load the model once and broadcast to all DP parallel ranks. We enabled broadcasting in two ways : either by using torch.broadcast or by using torch.all\_reduce distributed calls. This technique eliminates the hangs and improves the startup time significantly especially for the large models.
\\\\\\
\textbf{Dual checkpointing} : Checkpointing is used in deep learning model training to ensure minimal loss of work in case of training run failure or to resume training when compute resources are not available continuously. In checkpointing, all the necessary information (model parameters, optimizer states, training step etc.) required to resume training are stored. During checkpointing, failures can happen for multiple reasons such as filesystem I/O issues, resource relinquishment during checkpointing etc. Hence, having a single checkpoint does not help us in resuming training incase of a checkpointing failure. To overcome this problem, we implemented dual checkpointing feature, where two checkpoints are maintained instead of one, which ensures that there is always a valid checkpoint to resume the training. Consider a training scenario with dual checkpointing, where checkpointing is done at every 1000 steps. At step-1000, checkpointing is done at ckpt-1. At step-2000, checkpointing is done at ckpt-2. At step-3000, ckpt-1 is chosen for checkpointing as it is oldest of the two. Now, if a failure occurs while checkpointing to ckpt-1, it will not be an issue as ckpt-2 is valid and training can be resumed from that.
If we use single checkpoint, we cannot resume the training run if a checkpointing failure occurs. 
\\\\
\textbf{Persistent model checkpointing} : Dual checkpointing ensures that there is always a valid checkpoint to resume training regardless of how the training is progressing. But there can be issues with training itself like gradient explosion, data corruption leading to divergence etc. So, inorder to track back to a good training regime, we implemented persistent model only checkpointing feature. In this feature, we checkpoint only model parameters, which has significantly less memory footprint when compared to full checkpoint. For BF16 mixed precision training with AdamW optimizer, model only checkpoint takes 8x less memory than the full checkpoint. Training can be restarted from just the model parameters expect that optimizer states have to be initialized from scratch. We found that doing this does not alter the training in any significant manner. Consider a training scenario where the model is trained for 15000 steps and after 10000 steps the loss curve and gradient values are diverging. Because we store model only checkpoints at every 1000 steps, we can restart the model from step-10000 model only checkpoint with default optimizer states.
\\\\
\textbf{DP-Scattered model checkpointing} : In DP, the model is replicated across all ranks and model checkpointing is done by the rank corresponding to the first dp index i.e, the first rank. But when we train large models with model parallelism, following a similar method where ranks corresponding to first dp index write the model parallel shards will result in concentrated writes from a small set of nodes. In DP-Scattered model checkpointing, we spread the load where model parallel shard $m$ is written by data parallel index $d$, where ($d = m\%DP$). For example, if we train a model on 12 nodes with 12-way model parallelism, dp index is the node id as each node has 12 GPU tiles. If only dp\_index=0 writes, then all the model shards are written from a single node. But with DP-Scatterd model checkpointing, model shard $m$ is written by node $m$.
\\\\
\textbf{Hard node failure handling} : A training run is launched on a certain number of nodes. A failure can occur in a particular node for a variety of reasons like ping failure, segmentation fault, os error etc. When that happens, the training run has to restarted by excluding the node that failed. In hard node failure handling feature, we do this restarting automatically by launching the training run with some extra buffer nodes and restart the run by replacing the failed node with one of the buffer nodes.
\\\\
\textbf{Soft node failure handling} : In a hard node failure, the training run exits immediately after a node failure. But in a soft node failure, the training run continues to run but produces local NaN values in the soft failed node. If it is not detected, it will lead to NaN weights and contaminates checkpoints with NaNs. To avoid this, we check local loss and gradients for NaN in each rank, and if it occurs we mark the node corresponding to the NaN rank and exit the training run. We then automatically relaunch the training run by replacing the failed soft node with one of the buffer nodes.

\section{Conclusion}
We demonstrated the capability of Aurora super computer by pretraining MoE large language models on thousands of Intel GPUs for trillions of tokens. On the scale front, we pushed the model and compute scale to 220 Billion parameters and 12288 GPU tiles respectively and observed scaling efficiency of around 90\% for the Mula-220B-A10B model when going from 384 to 12288 GPU tiles.

\section*{Acknowledgments}
We thank Argonne National Labs for providing access to the Aurora Super Computer and continued support through out the duration of the work.

\printbibliography

\end{document}